\documentclass[runningheads]{llncs}
\usepackage[T1]{fontenc}
\usepackage{graphicx,verbatim}
\usepackage{subcaption}
\usepackage{booktabs}
\usepackage{multirow}
\usepackage{makecell}
\usepackage{amsmath,amssymb,amsfonts}
\usepackage[misc]{ifsym}
\begin{document}
\title{Landmarks Are Alike Yet Distinct: Harnessing Similarity and Individuality for One-Shot Medical Landmark Detection}

\titlerunning{Landmarks are alike yet distinct: Harnessing similarity and individuality}
%

\author{
    Xu He\inst{1,2}\thanks{X. He and Z. Huang contribute equally to this work.}  \and
    Zhen Huang\inst{3,4} \and
    Qingsong Yao\inst{5} \and
    Xiaoqian Zhou\inst{1,2} \and
    S. Kevin Zhou\inst{1,2}\textsuperscript{\Letter}
}
\authorrunning{X. He et al.}
\institute{\
School of Biomedical Engineering, Division of Life Sciences and Medicine, University of Science and Technology of China, Hefei, Anhui, 230026, P.R.China \\
\email{skevinzhou@ustc.edu.cn}\and
Suzhou Institute for Advanced Research, University of  Science and Technology of China, Suzhou, Jiangsu, 215123, P.R.China \and
School of Computer Science and Technology, University of Science and Technology of China, Hefei, Anhui, 230026, P.R.China \and 
School of Information Science and Technology, Eastern Institute of Technology (EIT), Ningbo, Zhejiang, 315200, P.R.China \and
Stanford University, Palo Alto, California, 94305, United States
}
    
\maketitle              

\begin{abstract}
Landmark detection plays a crucial role in medical imaging applications such as disease diagnosis, bone age estimation, and therapy planning. However, training models for detecting multiple landmarks simultaneously often encounters the "seesaw phenomenon", where improvements in detecting certain landmarks lead to declines in detecting others. Yet, training a separate model for each landmark increases memory usage and computational overhead. To address these challenges, we propose a novel approach based on the belief that "landmarks are distinct" by training models with pseudo-labels and template data updated continuously during the training process, where each model is dedicated to detecting a single landmark to achieve high accuracy. Furthermore, grounded on the belief that "landmarks are also alike", we introduce an adapter-based fusion model, combining shared weights with landmark-specific weights, to efficiently share model parameters while allowing flexible adaptation to individual landmarks. This approach not only significantly reduces memory and computational resource requirements but also effectively mitigates the seesaw phenomenon in multi-landmark training. Experimental results on publicly available medical image datasets demonstrate that the single-landmark models significantly outperform traditional multi-point joint training models in detecting individual landmarks. Although our adapter-based fusion model shows slightly lower performance compared to the combined results of all single-landmark models, it still surpasses the current state-of-the-art methods while achieving a notable improvement in resource efficiency. 


\end{abstract}

\section{Introduction}

Accurate medical landmark detection (MLD) has widespread applications in clinical settings, such as disease diagnosis~\cite{head,DiseaseDiag2,pele}, bone age estimation~\cite{boneAge}, and therapy planning~\cite{therapyPlan1,therapyPlan2}. It also supports various downstream tasks, such as segmentation~\cite{segment,segment2}, image reconstruction~\cite{reconstruction}, and image registration~\cite{registration}. With the rapid development of deep learning~\cite{deepLearning,deepLearning2}, many neural network-based models have been proposed for MLD. For instance, \cite{celda} employs a prototypical network for MLD by comparing image features with landmark prototypes, while \cite{hybrid} incorporates dynamic sparse attention into a hybrid Transformer-CNN architecture. However, despite the superior performance of these methods, they generally rely on a large amount of labeled data, which poses a significant challenge due to the time-consuming and labor-intensive annotation process. To address this issue, some studies have proposed {\it one-shot learning} methods, which use a single annotated medical image for landmark detection~\cite{yao2021one,miao2024fm,2stage}. 

MLD is typically formulated as a multi-label task, where traditional methods often train a single model to detect all landmarks, sharing the same network weights and thus ignoring the individuality of different landmarks. Since different landmarks may have distinct features and local variations, training them in the same network may lead to the so-called "seesaw phenomenon"~\cite{ple}, that is, improving the detection of certain landmarks could degrade the performance of others. In multi-task learning~\cite{multiTaskSurvey}, some studies have used hard parameter sharing~\cite{hardParaShare} to facilitate joint learning. \cite{MoE} introduced the Mixture of Experts (MoE) model, which shares some experts at the bottom layers and combines them through a gating network. \cite{ple} proposed Progressive Layered Extraction (PLE) to alleviate the seesaw phenomenon. Inspired by these approaches, we introduce the concept of multi-task learning into landmark detection and propose a novel single-landmark approach (SLA) to harness the {\it landmark individuality} and improve the accuracy of one-shot landmark detection.

We use the Cascade Comparing to Detect (CC2D) framework~\cite{yao2021one}, a pioneering one-shot MLD method, as the foundation and propose the CC2D-SLA method, which eliminates the interdependencies between different landmarks by training a separate model for each landmark, thereby addressing the seesaw phenomenon at its core. Our experiments demonstrate that CC2D-SLA brings improved MLD accuracy over CC2D. To further enhance the detection accuracy of each landmark, we leverage the idea of augmented template data (ATD)~\cite{augment}, which leads to an improved SLA method called CC2D-SLA-ATD. 

However, both CC2D-SLA and CC2D-SLA-ATD require training a separate model for each landmark, which leads to a substantial computational overhead. To address the issue of resource inefficiency and redundancy bought by multiple single-landmark models and further mitigate the seesaw phenomenon, we introduce an adapter to harness the {\it landmark similarity and individuality}. By combining shared weights with landmark-specific weights, we enable the model to learn the features of all landmarks through a single end-to-end model. Ultimately, this leads to the proposed CC2D-SLA-ATD-Adapter method, which not only reduces computational and memory overhead but also maintains high precision in landmark detection performance.

We evaluate our models on the ISBI 2015 Challenge dataset~\cite{head}. Experimental results show that the single-landmark models significantly outperform traditional multi-point joint training models in landmark detection. Although our adapter-based fusion model (CC2D-SLA-ATD-Adapter) slightly underperforms compared to the best results from combining all single-landmark models, it still outperforms the current state-of-the-art (SOTA), demonstrating the potential of our method for medical imaging applications.

\section{Method}
\begin{figure}[!t]
\centerline{\includegraphics[width=0.9\columnwidth]{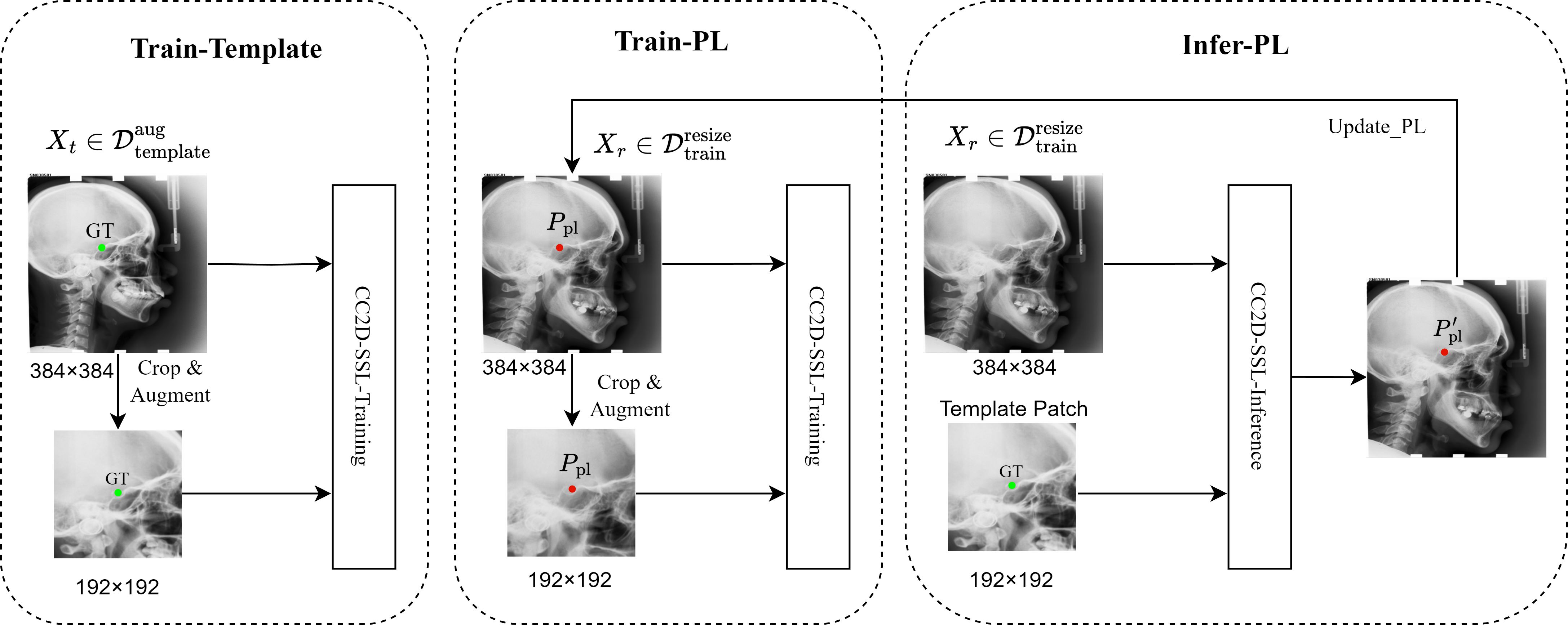}}
\caption{Training framework of CC2D-SLA-ATD, which consists of three stages at each epoch: Train-Template, Train-PL, and Infer-PL. CC2D-SLA training, on the other hand, is composed of only the Train-PL and Infer-PL stages.}
\label{fig:CC2D-SLA-ATD}
\end{figure}

Below, we first introduce CC2D-SLA and its improved variant, CC2D-SLA-ATD. We then describe how to integrate the adapter. Finally, we present the architecture of our CC2D-SLA-ATD-Adapter. Fig.~\ref{fig:CC2D-SLA-ATD} shows the training framework.

\subsection{CC2D-SLA}
The training of CC2D-SLA in each epoch consists of two stages: 1) Train-PL, in which the model is trained using pseudo-labels (PLs), and 2) Infer-PL, in which the pseudo-labels are inferred and updated.

\noindent\textbf{Train-PL Stage.} 
Let $\mathcal{D}_{\text{train}}$ be the training dataset. We first resize all images to $384\times384$, forming the resized set $\mathcal{D}_{\text{train}}^\text{resize}$. Initially, each image $X_r \in \mathcal{D}_{\text{train}}^\text{resize}$ is assigned a random pseudo-label $P_{pl} = (x_{pl}, y_{pl})$. As training proceeds, these pseudo-labels are updated at the end of each epoch.
During the Train-PL stage, for each image $X_r$, we first crop a patch centered at $P_{pl}$ and then apply data augmentation to obtain $X_p$. Both $X_r$ and $X_p$ are subsequently used as inputs to the CC2D-SSL framework in its training stage, referred to as CC2D-SSL-Training. Note that CC2D-SSL-Training encompasses the entire training pipeline of CC2D-SSL after receiving the image inputs, and is distinct from our CC2D-SLA approach. Here, we simply replace the typical input of CC2D-SSL-Training with the pair $(X_r, X_p)$ to train on the PL-based patches.

\noindent\textbf{Infer-PL Stage}. 
After one pass through the training set, the model enters the Infer-PL stage to update the pseudo-labels. We denote CC2D-SSL-Inference as the complete inference pipeline of the CC2D-SSL framework. 
Specifically, for a given landmark ID $k$, a template patch $X_{tp}$ is cropped from the template image around the ground-truth location of the $k$-th landmark. Next, every image $X_r \in \mathcal{D}_{\text{train}}^\text{resize}$ serves as a query image. We feed $(X_{tp}, X_r)$ into the CC2D-SLA-Inference module to infer the new pseudo-label $P'_{pl}$ for the $k$-th landmark on $X_r$. Thus, all pseudo-labels in $\mathcal{D}_{\text{train}}^\text{resize}$ are updated at the end of the epoch.

\noindent\textbf{CC2D-SLA-ATD.} 
In CC2D-SLA, training relies on pseudo-labels generated from a single template image. To better utilize the template data, we perform data augmentation on the template, following the procedure in FM-OSD~\cite{miao2024fm}. Specifically, we apply random shifting, rotation, and scaling to produce 500 augmented versions of the template, forming the dataset $\mathcal{D}_{\text{template}}^\text{aug}$. 

As illustrated in Fig.~\ref{fig:CC2D-SLA-ATD}, for each augmented template image $X_t \in \mathcal{D}_{\text{template}}^\text{aug}$, we crop a patch centered at its ground-truth location and apply additional data augmentation to obtain $X_{tp}$. We then feed $X_t$ and $X_{tp}$ into the CC2D-SSL-Training pipeline. We refer to this procedure as the Train-Template stage, which is prepended to CC2D-SLA to get CC2D-SLA-ATD.

\subsection{Adapter Integration for Multi-Landmark Training}
\label{sec:adapter_method}
To address the seesaw phenomenon and the resource inefficiency of multiple single-landmark models, we introduce adapter layers into our network. By incorporating adapters, all landmarks can be jointly trained using shared and landmark-specific weights, while ensuring only minor performance degradation.

\noindent\textbf{Pre-Adapter Feature Extraction.} 
As shown in the upper-left part of Fig.~\ref{fig:CC2D-SLA-ATD-Adapter}, without adapters, a feature map $F_i$ of shape $H \times W \times C$ is transformed by a convolutional layer $Conv$, yielding $F_{i+1} \in \mathbb{R}^{H \times W \times C'}$:
$
    F_{i+1} = Conv(F_i).
$

\noindent\textbf{Adapter-Incorporated Layers.} 
After integrating adapters, each landmark $k$ has its own dedicated convolution $Conv\text{-}A_k$ alongside the shared convolution $Conv$. The primary difference lies in the output channel dimension: $Conv\text{-}A_k$ produces a feature map $F_{i+1}^{A_k} \in \mathbb{R}^{H \times W \times C_A}$, where $C_A$ (e.g., 16) is typically much smaller than $C'$. Formally,
$
F_{i+1}^{A_k} = Conv\text{-}A_k(F_i).
$
We then concatenate $F_{i+1}$ and $F_{i+1}^{A_k}$ along the channel dimension to obtain 
\begin{align}
    F'_{i+1} = \text{Concat}\bigl(F_{i+1}, F_{i+1}^{A_k}\bigr) = \text{Concat}\Bigl(Conv(F_i),\, Conv\text{-}A_k(F_i)\Bigr),\label{eq:ConvAdapter}
\end{align}

where $F'_{i+1} \in \mathbb{R}^{H \times W \times (C' + C_A)}$. Equation \eqref{eq:ConvAdapter} is simplified as $F'_{i+1}=C\text{-}A_k(F_i)$. Consequently, the subsequent layer’s input channel size is adjusted to $C' + C_A$. 

In this scheme, $Conv$ captures shared features across all landmarks, while $Conv\text{-}A_k$ learns landmark-specific features, activated only during training or inference of landmark $k$. This design alleviates the seesaw phenomenon, and enables landmark-specific learning without affecting others.

Adapters allow joint training of multiple landmarks, reducing the need for separate models for each, thus improving memory and computational efficiency. This shared-plus-specific design balances accuracy and resource usage, maintaining performance when scaling to multiple landmarks.

\subsection{CC2D-SLA-ATD-Adapter}

By enhancing CC2D-SLA-ATD with an adapter, we achieve CC2D-SLA-ATD-Adapter, enabling multi-landmark training while mitigating the seesaw phenomenon and improving overall performance.

\begin{figure*}[!t]
\centerline{\includegraphics[width=0.9\textwidth]{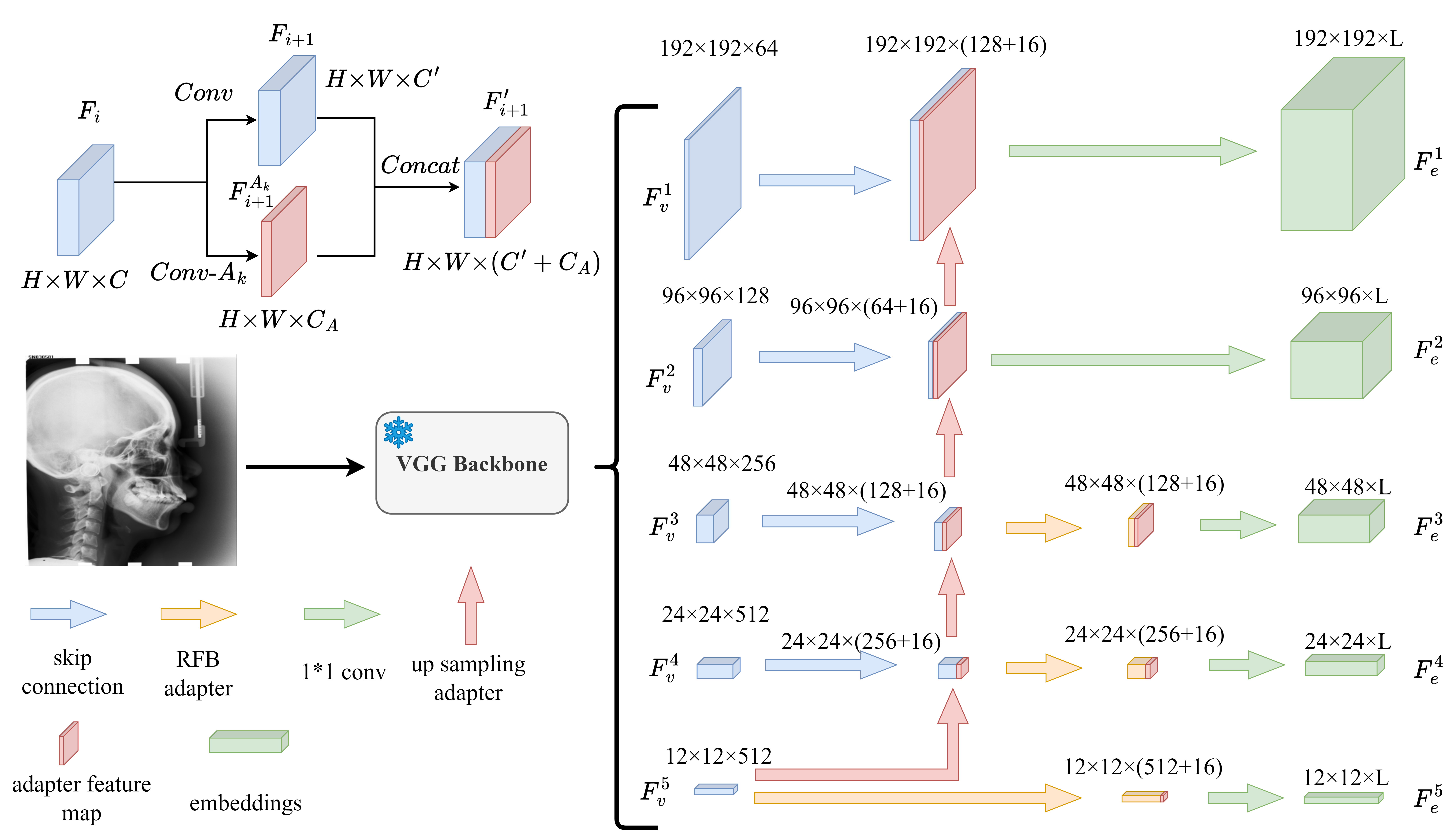}}
\caption{Network architecture of CC2D-SLA-ATD-Adapter. 
The upper-left part of the figure shows an illustration of the adapter-based feature map transformation.}
\label{fig:CC2D-SLA-ATD-Adapter}
\end{figure*}

As illustrated in Fig.~\ref{fig:CC2D-SLA-ATD-Adapter}, for an input image $X$ of size $384\times384$, we use a pretrained VGG~\cite{vgg} network to extract five layers of feature maps:
$
    F_v = \{F_v^1, F_v^2, \dots, F_v^5\} = \mathrm{VGG}(X).
$
Here, $F_v^5$ is the deepest layer. We adopt the same VGG19 model~\cite{vgg} pretrained on ImageNet~\cite{deng2009imagenet} as in CC2D~\cite{yao2021one}, and keep its weights frozen throughout training. Following~\cite{unet}, 
each upsampled feature map is concatenated with the corresponding lower-level feature map. Specifically,
\begin{equation}
    F_{\text{concat}}^i = \mathrm{Concat}\Bigl(\mathrm{Up}\bigl(F_c^{i+1}\bigr), \, F_v^i\Bigr),
    \quad i = 1,2,3,4.
\end{equation}
Here, $F_c$ is the feature map obtained by applying the adapter-based convolution:
\begin{equation}
    F_c^i = C\text{-}A_k\bigl(F_{\text{concat}}^i\bigr),
    \quad i = 1,2,3,4,
\end{equation}
We set $F_c^5 = F_v^5$ to represent the deepest layer (where no upsampling occurs).

In CC2D~\cite{yao2021one}, the third, fourth, and fifth layers employ Receptive Field Block (RFB) modules~\cite{rfb} to enlarge the receptive field. We also integrate adapters into the RFB modules by replacing the $Conv$ and $Conv\text{-}A_k$ modules with $RFB$ and $RFB\text{-}A_k$, then concatenate their outputs. Formally,
\begin{align}
    F_{RFB}^i = R\text{-}A_k\bigl(F_c^i\bigr) = \mathrm{Concat}\Bigl(\mathrm{RFB}(F_c^i), \, \mathrm{RFB\text{-}A_k}(F_c^i)\Bigr),
    \quad i=3,4,5,
\end{align}

where $F_{RFB}^i$ is the resulting feature map after applying the RFB with adapters.

Finally, each feature map is passed through a $1\times1$ convolution to yield a fixed-dimensional embedding $F_e^i$. In particular,
\begin{equation}
    F_e^i =
    \begin{cases}
        \mathrm{Conv}_{1\times1}\bigl(F_c^i\bigr), & \text{if } i = 1, 2, \\
        \mathrm{Conv}_{1\times1}\bigl(F_{RFB}^i\bigr), & \text{if } i = 3, 4, 5.
    \end{cases}
\end{equation}

\section{Experiments}
\subsection{Settings}
\noindent\textbf{Dataset.} 
For this study, we use the widely recognized IEEE ISBI 2015 Challenge dataset~\cite{head}, which contains 400 radiographs annotated with 19 landmarks by two expert clinicians. The average of their annotations serves as the ground truth. The images are 1935 × 2400 pixels with a 0.1mm pixel spacing. The dataset is split into 150 training and 250 testing images. One image is selected as the template, and the others are treated as unlabeled data for model training.

\noindent\textbf{Evaluation Metrics.} 
We evaluate the model performance using two common metrics: Mean Radial Error (MRE) and Successful Detection Rate (SDR). MRE calculates the average Euclidean distance between predicted landmarks and ground truth. SDR measures the proportion of landmarks detected within various thresholds (2mm, 2.5mm, 3mm, and 4mm) from the ground truth. These metrics are widely used in previous studies on landmark detection~\cite{yao2021one,miao2024fm,uod}.

\noindent\textbf{Implementation Details.} 
All experiments are conducted using PyTorch on an NVIDIA RTX 3090 GPU with a learning rate of 0.0001, the Adam optimizer, a batch size of 8, and 300 epochs. In CC2D-SLA-ATD (or C-ATD in short), 19 models are trained, each dedicated to a single landmark. For CC2D-SLA-ATD-Adapter (or C-Adapter in short), we introduce 19 adapters, each with an output channel size of 16. A frozen VGG19 network serves as the feature extractor.

\subsection{Performance Comparison}

\begin{table}[t]
\centering
\fontsize{8pt}{9pt} \selectfont
\caption{Performance comparison of different methods on the Head~\cite{head} dataset.}
\begin{tabular}{lcccccc}
\toprule
\multirow{2}{*}{Method} & \multirow{2}{*}{\makecell{Model \\ Count}} & \multirow{2}{*}{\makecell{MRE(↓) \\ (mm)}}  & \multicolumn{4}{c}{SDR(↑)(\%)} \\ 
\cmidrule(lr){4-7}  
 & &  & 2mm & 2.5mm & 3mm & 4mm \\
\midrule
SAM~\cite{yansam} & 1 & 2.56 & 54.11 & 63.66 & 70.25 & 80.84 \\
UOD~\cite{uod} & 1 & 2.43 & 51.14 & 62.37 & 74.40 & 86.49 \\
CC2D~\cite{yao2021one} & 1 & 2.04 & 62.46 & 71.62 & 80.00 & 89.45 \\
FM-OSD(coarse)~\cite{miao2024fm} & 1 & 1.93 & 63.60 & 75.43 & 83.03 & 91.94 \\
FM-OSD(fine)~\cite{miao2024fm} & 2 & 1.82 & 67.35 & 77.92 & 84.59 & 91.92 \\
CC2D-SLA(ours) & 19 & 1.82 & 69.73 & 76.69 & 84.04 & \textbf{92.17}\\
C-ATD(ours) & 19 & \textbf{1.79} & \textbf{72.02} & \textbf{78.02} & \textbf{84.72} & 92.00\\
C-Adapter(ours) & 1 & 1.96 & 67.83 & 75.33 & 81.89 & 90.82 \\
C-F2(ours) & 3 & 1.83 & 70.48 & 77.35 & 83.96 & 91.43\\
\bottomrule
\end{tabular}
\label{tab:performance_comparison}
\end{table}

As shown in Table~\ref{tab:performance_comparison}, we compare our proposed models with several SOTA methods, including SAM~\cite{yansam}, UOD~\cite{uod}, CC2D~\cite{yao2021one}, and the two stages of FM-OSD~\cite{miao2024fm}. We also implement CC2D-SLA-ATD-Adapter-F2 (or C-F2 in short) , which uses our C-Adapter model as the coarse stage of FM-OSD, followed by the fine stage of FM-OSD to perform landmark detection on high-resolution medical images. Note that all models except C-F2 are evaluated on low-resolution images (384$\times$384). We report the MRE and SDR results for these methods on the ISBI 2015 Challenge dataset and compare the number of models used.

It is evident that our C-ATD model achieves the best performance, with a 2mm SDR of 72.02\% and an MRE of 1.79mm, significantly surpassing previous SOTA methods. This confirms that training individual landmark models effectively enhances detection accuracy, though it requires 19 separate models. To improve efficiency, we introduce adapters, allowing for the fusion of all landmarks into a single model. While this results in a slight performance drop compared to C-ATD, it still performs similarly to FM-OSD's coarse stage. Furthermore, applying FM-OSD's fine stage for high-resolution inference boosts the performance of C-Adapter, achieving a 2mm SDR of 70.48\%. {\it This not only exceeds the previous SOTA methods but also sets a new benchmark for MLD performance}.

\begin{figure}[t]
    \centering
    \begin{minipage}{0.62\textwidth}
        \centering
        \fontsize{8pt}{9pt} \selectfont
        \captionof{table}{The performances of our methods with different channel sizes.}
        \begin{tabular}{c|cccccc}
        \toprule
        \multirow{2}{*}{Para.} & \multirow{2}{*}{Value} & \multirow{2}{*}{\makecell{MRE(↓) \\ (mm)}}  & \multicolumn{4}{c}{SDR(↑)(\%)} \\ 
        \cline{4-7}
         & &  & 2mm & 2.5mm & 3mm & 4mm \\
        \midrule
        \multirow{5}{*}{$C_A$} & 0 & 2.30 & 63.92 & 72.15 & 79.68 & 89.09 \\
        & 4  & 1.95 & 67.64 & \textbf{75.92} & 82.61 & \textbf{91.07} \\
        & 8  & 1.95 & 67.07 & 74.53 & 81.18 & 90.48 \\
        & 16 & 1.96 & 67.83 & 75.33 & 81.89 & 90.82 \\
        & 32 & \textbf{1.91} & \textbf{68.21} & 75.64 & \textbf{82.67} & 90.99 \\
        \midrule
        \multirow{2}{*}{\makecell{CC2D's \\ channels}} & +16 & 2.06 & 62.00 & 70.38 & 79.16 & 88.76\\
        & +16×19 & \textbf{2.02} & \textbf{63.68} & \textbf{72.32} & \textbf{80.38} & \textbf{89.68}\\
        \bottomrule
        \end{tabular}
        \label{tab:ablation_study}
    \end{minipage}%
    \hfill
    \begin{minipage}{0.37\textwidth}
        \centering
        \includegraphics[width=\linewidth]{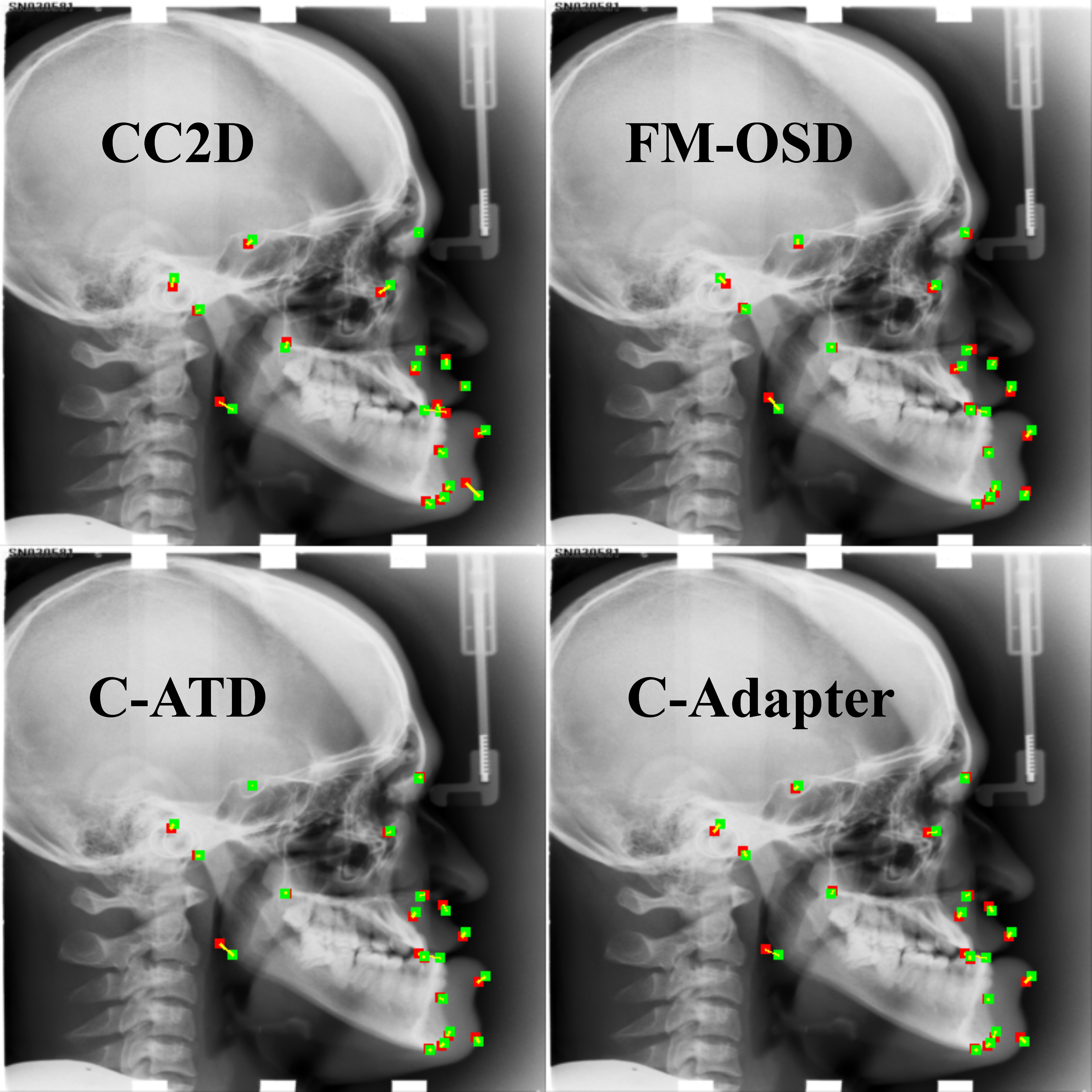}
        \caption{Visualizations of the prediction results from different methods. The landmarks in red and green represent the predictions and ground truths.}
        \label{fig:prediction-image}
    \end{minipage}
\end{figure}

Fig.~\ref{fig:prediction-image} presents the predicted landmark detection results from different methods on the dataset. As shown in the figure, CC2D has the lowest prediction accuracy among the methods displayed, while FM-OSD achieves relatively better accuracy. The prediction accuracy of C-Adapter is comparable to that of FM-OSD. And the C-ATD model shows the best prediction accuracy overall.

\begin{table}[t]
\centering
\fontsize{8pt}{8pt} \selectfont
\caption{Performance of different methods on the 19 landmarks. The landmark indices correspond to the positions described in~\cite{headindex}. The MRE is measured in millimeters, and the SDR refers to the 2mm SDR, with units in percentage.}
\begin{tabular}{ccccccccc}
\toprule
\multirow{2}{*}{Landmark} & \multicolumn{2}{c}{CC2D} & \multicolumn{2}{c}{FM-OSD} & \multicolumn{2}{c}{C-ATD} & \multicolumn{2}{c}{C-F2} \\
\cmidrule(lr){2-3}  
\cmidrule(lr){4-5}
\cmidrule(lr){6-7}
\cmidrule(lr){8-9}
 & MRE & SDR(2mm) & MRE & SDR(2mm) & MRE & SDR(2mm) & MRE & SDR(2mm)\\
\midrule
1  & 1.35 & 85.2 & 1.55 & 84.0 & \textbf{0.98} & \textbf{97.6} & 1.26 & 92.0 \\
2  & 1.60 & 74.0 & \textbf{1.49} & 73.2 & 1.51 & \textbf{78.0} & 1.54 & 73.6 \\
3  & 1.58 & 72.4 & 1.66 & 68.4 & \textbf{1.37} & \textbf{84.4} & 1.46 & 78.4 \\
4  & 1.93 & 66.4 & 2.28 & 55.6 & \textbf{1.80} & \textbf{70.8} & 2.00 & 67.2 \\
5  & 1.86 & 62.8 & 1.72 & 65.6 & \textbf{1.54} & \textbf{76.0} & 1.61 & 75.2 \\
6  & 2.50 & \textbf{50.0} & \textbf{2.41} & 48.4 & 2.47 & 49.2 & 2.43 & 48.4 \\
7  & 1.42 & 81.6 & 1.05 & 88.0 & 0.96 & \textbf{94.0} & \textbf{0.94} & 93.6 \\
8  & 1.46 & 78.4 & \textbf{0.99} & 91.6 & 1.10 & \textbf{93.2} & 1.93 & 89.6 \\
9  & 1.08 & 88.8 & \textbf{0.83} & 94.8 & 0.84 & \textbf{96.8} & 0.84 & 95.2 \\
10 & 4.19 & 20.4 & 3.23 & 33.6 & 3.59 & 24.8 & \textbf{3.17} & \textbf{37.6} \\
11 & 2.58 & 42.8 & \textbf{2.44} & \textbf{52.0} & 2.85 & 48.8 & 2.66 & 48.8 \\
12 & 2.60 & 48.8 & 1.91 & 68.0 & \textbf{1.43} & \textbf{86.4} & 1.50 & 80.8 \\
13 & 1.63 & 70.0 & 1.60 & 68.4 & 1.56 & 78.4 & \textbf{1.52} & \textbf{79.2} \\
14 & 1.69 & 71.2 & \textbf{1.43} & 80.0 & 1.54 & 79.6 & 1.47 & \textbf{80.0} \\
15 & 1.73 & 65.6 & \textbf{1.69} & \textbf{68.8} & 2.50 & 50.0 & 1.89 & 59.6 \\
16 & 3.54 & 26.4 & 2.61 & 47.2 & \textbf{2.52} & \textbf{51.2} & 2.56 & 48.8 \\
17 & 1.73 & 67.6 & 1.55 & 74.8 & \textbf{1.24} & \textbf{84.0} & 1.51 & 77.6 \\
18 & 1.77 & 65.6 & \textbf{1.67} & \textbf{67.6} & 2.00 & 63.2 & 1.82 & 66.4 \\
19 & 2.44 & 48.8 & 2.49 & 49.6 & \textbf{2.25} & \textbf{62.0} & 2.75 & 47.2 \\
Mean & 2.04 & 62.5 & 1.82 & 67.4 & \textbf{1.79} & \textbf{72.0} & 1.83 & 70.5 \\
\bottomrule
\end{tabular}
\label{tab:single_landmark}
\end{table}

As shown in Table~\ref{tab:single_landmark}, we provide the MRE and SDR results for each of the 19 landmarks across different methods. C-ATD achieves the best performance on most landmarks, with significant improvements for landmarks 1, 3, 5, 7, 12, 17, and 19, highlighting the effectiveness of the single-landmark approach. However, its performance is less optimal for landmarks 10, 15, and 18, suggesting that certain landmarks require more domain-specific knowledge. To address this, we introduce adapters, allowing landmarks to learn through shared weights as well as landmark-specific weights. After testing, this adaptation leads to a more balanced performance across all landmarks.

\subsection{Ablation Study}

As shown in the upper half of Table~\ref{tab:ablation_study}, we investigate the effect of varying the output channel size, denoted as $C_A$, for each adapter in the C-Adapter model. We set $C_A$ to 0, 4, 8, 16, and 32. When no adapter is used ($C_A = 0$), the model's performance significantly degrades. However, with adapters, performance remains stable across different $C_A$ values, suggesting that the adapters help the model recognize feature patterns, while shared weights handle main feature extraction.

In the lower half of Table~\ref{tab:ablation_study}, we explore the impact of increasing the number of channels in all convolution layers of the decoder in the original CC2D model, in order to examine how weight scaling affects performance. Specifically, we increased the number of channels by 16 and by 16×19. While increasing weights improves performance slightly, the gain is small, indicating that C-Adapter's performance enhancement is not primarily due to increased model weights.

\section{Conclusion}
In this paper, we present a progression from exploiting each landmark's individuality—through single-landmark training—to utilizing inter-landmark similarity by incorporating adapters into a unified model. This two-fold approach demonstrates a feasible and effective strategy for improving MLD accuracy. The proposed C-Adapter represents an initial endeavor toward jointly learning multiple landmarks via shared and landmark-specific weights. However, further investigation is needed into more advanced methods of inter-landmark collaboration to simultaneously enhance performance for all landmarks. We believe that continued exploration of this balance between individuality and similarity will yield more robust, efficient, and accurate solutions for one-shot MLD.

%
\bibliographystyle{splncs04}
\bibliography{myref}
%




\end{document}